\documentclass[11pt, journal]{IEEEtran} 
\usepackage{graphicx}
\usepackage{listings}
\usepackage{amsmath,amssymb,amsthm,mathrsfs,amsfonts,dsfont}
\usepackage{rotating}
\usepackage{authblk}
\usepackage{hyperref}

\newcommand{\bx}{\mathbf{x}}
\DeclareMathOperator*{\argmin}{arg\,min}

\makeatletter
\newcommand*{\defeq}{\mathrel{\rlap{%
                     \raisebox{0.3ex}{$\m@th\cdot$}}%
                     \raisebox{-0.3ex}{$\m@th\cdot$}}%
                     =}
\makeatother

\ifCLASSINFOpdf
\else
\fi
\begin{document}
%
\title{Detecting Relative Anomaly}
%
%
%

\author{Richard Neuberg and~Yixin Shi
\thanks{R. Neuberg is with the Department of Statistics, Columbia University, New York, NY 10027, USA, e-mail: rn2325@columbia.edu.}
\thanks{Y. Shi is with Google, Inc.}
}

%
%

\markboth{
}%
{Shell \MakeLowercase{\textit{et al.}}: Bare Demo of IEEEtran.cls for Journals}
%



\maketitle

\begin{abstract}
System states that are anomalous from the perspective of a domain expert occur frequently in some anomaly detection problems. The performance of commonly used unsupervised anomaly detection methods may suffer in that setting, because they use frequency as a proxy for anomaly.  We propose a novel concept for anomaly detection, called \textit{relative anomaly detection}.  It is tailored to be robust towards anomalies that occur frequently, by taking into account their location relative to the most typical observations. The approaches we develop are computationally feasible even for large data sets, and they allow real-time detection.  We illustrate using data sets of potential scraping attempts and Wi-Fi channel utilization, both from Google, Inc.
\end{abstract}


%
\IEEEpeerreviewmaketitle

\section{Introduction}

Multivariate anomaly detection may be categorized broadly into supervised and unsupervised detection. In supervised anomaly detection, training data are labeled by domain experts as normal or anomalous, and a model is trained to classify future observations.  In unsupervised anomaly detection, which is the focus of this article, labels are not known, because labeling is too difficult or costly.  The goal is to approximately recover the missing expert judgements using empirical characteristics of the data.  The data themselves typically first undergo a feature selection and feature engineering process to devise informative covariates.  An unsupervised model can be evaluated by comparing its predictions with actual domain expert labels.  Potential applications include intrusion detection, fraud detection and process control.  

Frequency is commonly chosen as the target criterion for unsupervised anomaly detection.  The population definition of anomalous observations then is $\{\mathbf{x} :  f(\mathbf{x}) < \lambda\}$, where $f$ is the data generating density, and $\lambda$ is a user-selected threshold.  Methods that exactly or approximately fall under this paradigm are density estimators and the closely related nearest neighbor approaches, besides many others; for a review on commonly used anomaly detection methods, see \cite{chandola2009anomaly}.

However, frequency may not align well with expert judgements in some applications.  For example, scraping (the automated collection of information from websites) may occur frequently, but it nevertheless constitutes anomalous user behavior.  The performance of common approaches to unsupervised anomaly detection may suffer in the presence of such frequently occurring anomalies.

We propose a framework which we call relative anomaly detection to better handle cases where anomalies may occur frequently.  We use the term \textit{relative} to emphasize that in this framework the anomaly of an observation is determined by taking into account not only its own location and that of neighboring observations, but also the location of the most typical system states.  The underlying assumption in relative anomaly detection is that large clusters of high-density system states are indeed normal from an expert's perspective, and that observations that are far from these most typical system states are anomalous.  Such anomalies may occur frequently.

The rest of this paper is organized as follows. In Section~\ref{Graph}, we discuss the approach to anomaly detection of~\cite{moonesinghe2006outlier}, which is closely related to the PageRank algorithm~\cite{page1999pagerank}. We discuss the similarity graph of the observations in the training data set. We show connections with other approaches to anomaly detection, and discuss their shortcomings in the presence of anomalies that occur frequently. In Section~\ref{RelativeAnomaly}, we introduce two novel \textit{relative} anomaly detection approaches. In Section~\ref{Application}, we compare our approaches with that of \cite{moonesinghe2006outlier}, using data sets of potential scraping attempts and Wi-Fi usage from Google, Inc.  We conclude in Section~\ref{Conclusion}.

\section{Many approaches to anomaly detection target frequency criterion}\label{Graph}

In this section we show that the anomaly detection approach of \cite{moonesinghe2006outlier}, which is similar to the PageRank method \cite{page1999pagerank}, approximately targets the frequency criterion.  We show that it is also closely related to kernel density estimation and the nearest neighbor approach.  We begin by introducing the similarity graph, which will also serve as a basis for the relative anomaly detection approaches we develop in Section \ref{RelativeAnomaly}.

\subsection{Similarity graph}

The relationship between unlabeled observations in a data set may be described through a weighted similarity graph.  Observations form the nodes of the graph, and the weight of an edge expresses the similarity between two observations.  Two observations $\bx_i$ and $\bx_j$ are typically considered similar when their distance is small.  However, non-monotonic transformations can be useful with time series data, to take into account periodic behavior of the underlying system; for a reference on such transformations, see \cite[Chapter 4]{rasmussen2006gaussian}. A common monotonic transformation from distance $d(\bx_i, \bx_j)$ to similarity $s(\bx_i, \bx_j)$ uses the kernel function
\begin{equation}\label{rbfkernel}
	s(\bx_i, \bx_j) = \exp(-d(\bx_i,\bx_j)^2 / \gamma),
\end{equation}
which is symmetric in its arguments. The parameter $\gamma$ controls the degree of localization, meaning how far one observation can lie from another observation for the two to still be considered similar.  When $\gamma = \infty$, all observations are equally similar to $\bx_i$, and when $\gamma \downarrow 0$ only $\bx_i$ is similar to itself. More localization is needed when the data come from a complicated distribution.   The resulting matrix of similarities, $\mathbf{S}$, holds the edge weights in the similarity graph.  In methods that apply the ``kernel trick,'' such as the support vector machine and kernel principal components analysis, such a similarity matrix is called the kernel matrix.

Common choices for the distance between two real data points, $d(\bx_i,\bx_j)$, are Euclidean ($L_2$) and Manhattan ($L_1$) distance.  Both of these distance measures assume that each dimension of the data has been appropriately normalized.  Euclidean distance has the advantage of being rotation invariant, and the order of the resulting distances typically remains meaningful even in high dimension \cite{zimek2012survey}. Furthermore, data points often approximately lie in a lower-dimensional subspace; then Euclidean distance calculations are effectively carried out in the lower-dimensional subspace.  For very high-dimensional problems, Manhattan distance may be preferred over Euclidean distance \cite{aggarwal2001surprising}.  However, if the data truly cover the high-dimensional space, that means that the system components are barely correlated, even after feature selection and feature engineering.  Then a multivariate anomaly analysis may add only little value as compared to running separate univariate analyses.  If variables are measured on a nominal or ordinal scale, they may be converted into numerical data using dummy variables, or specialized distance measures for that scale level can be used; for a reference, see \cite[Chapter 14]{trevor2009elements}.

\subsection{A random walk approach, its relationships with other methods, and problems in the presence of frequently occurring anomalies}\label{vertexapproach}

The approach of \cite{moonesinghe2006outlier} proposes to take a random walk on the similarity graph, and to label an observation as anomalous when the stationary probability of the random walk at that observation is low.  For cases where the similarity matrix is not irreducible and aperiodic, random restarts are introduced in the random walk, like it was proposed as part of the PageRank algorithm \cite{page1999pagerank}.  In the case that the similarity matrix is irreducible and aperiodic, which we will assume in to following to keep technical discussions at a minimum, the matrix of transition probabilities in the graph is simply the similarity matrix normalized by row,
\begin{equation}
	\mathbf{P} = [\mathrm{diag}(\mathbf{S} \mathbf{1})]^{-1} \mathbf{S},
\end{equation}
where $\mathbf{1}$ is a column vector of ones.  The vector of stationary (unnormalized) probabilities, $\mathbf{p}$, follows from the stationarity condition $\mathbf{P}^\mathsf{T} \mathbf{p} = \mathbf{p}$ as the dominant left-eigenvector of $\mathbf{P}$ by the Perron--Frobenius theorem; see \cite{isaacson1976markov} for a reference. 

We now show that the approach of \cite{moonesinghe2006outlier} is closely related to both a density-based and a distance-based approach to anomaly detection.  
To see the connection with density-based anomaly detection, consider the case when $\mathbf{S}$ is symmetric; then the dominant left-eigenvector of $\mathbf{P} = [\mathrm{diag}(\mathbf{S} \mathbf{1})]^{-1} \mathbf{S}$ is, up to scaling, $\mathbf{S} \mathbf{1}$.  This follows from plugging in $\mathbf{S} \mathbf{1}$ for $\mathbf{p}$ in $\mathbf{P}^\mathsf{T} \mathbf{p} = \mathbf{p}$, and using that $\mathbf{P}^\mathsf{T} = \mathbf{S}^\mathsf{T} [\mathrm{diag}(\mathbf{S} \mathbf{1})]^{-1}$, with $\mathbf{S}^\mathsf{T} = \mathbf{S}$, which yields the true statement $\mathbf{S} [\mathrm{diag}(\mathbf{S} \mathbf{1})]^{-1} \mathbf{S} \mathbf{1} = \mathbf{S} \mathbf{1}$.  We see that the stationary probability at observation $\bx_i$ is proportional to its (weighted) vertex degree $\mathrm{VD}(\bx_i)$ in the similarity graph, where
\begin{equation}\label{vddegree}
	\mathrm{VD}(\bx_i) \defeq (\mathbf{S} \mathbf{1})_i = \sum_{j=1}^n s(\bx_i, \bx_j).
\end{equation}
Expression (\ref{vddegree}) is proportional to a kernel density estimate with Gaussian kernel, whose kernel covariance matrix is diagonal with all diagonal elements equaling $\gamma/2$.  As a density estimate, $\mathrm{VD}(\bx_i)$ is typically mis-specified, because the kernel matrix is not tuned to fit the particular data generating process. This may actually be desired in anomaly detection problems where a low density observation close to a very typical system state does not make for an interesting anomaly. However, the close connection with kernel density estimation suggests that if anomalous system states occur too frequently, they may not be labeled correctly as anomalies, even if they are for from the most typical system states.

To also see the connection with distance-based anomaly detection, consider a directed $k$ nearest neighbor graph instead of a fully connected similarity graph. Here the $(i,j)$th element of $\mathbf{S}$ takes value $s(\bx_i, \bx_j)$ if $\bx_j$ is in the set $\mathcal{N}_k(\bx_i)$, which contains the $k$ nearest neighbors of $\bx_i$, and it is zero otherwise.  The resulting similarity matrix is a sparse approximation of the full similarity matrix.  The additional tuning parameter $k$ controls the degree of localization.  Localization via the $k$ nearest neighbor graph is also used in spectral clustering, manifold learning, and local multidimensional scaling; for a reference, see \cite[Chapter 14]{trevor2009elements}. Consider a linear expansion of the radial kernel function, defined in Equation~(\ref{rbfkernel}), around some distance level $v > 0$.  Then $\mathrm{VD}(\bx_i)$ is approximately an affine decreasing function of the average distance to the $k$ nearest neighbors:
\begin{align}  
	&\mathrm{VD}^\mathrm{approx}(\bx_i) = 
	  k \exp(-v^2 / \gamma) (1 + 2  v^2  / \gamma)  \\
	& \qquad\quad - 2 (v \exp(-v^2 / \gamma))/ \gamma \sum_{j : \bx_j \in \mathcal{N}_k(\bx_i)}   d(\bx_i,\bx_j). \notag
\end{align}
This effectively eliminates the dependency on the kernel parameter $\gamma$.  Using the average distance to the $k$ nearest neighbors as a measure of anomaly was suggested in both \cite{eskin2002geometric} and \cite{angiulli2006distance}.  However, for relative anomalies, the average distance to the $k$ nearest neighbors can be small, and what is an anomalous system state may not be considered anomalous by the anomaly detection model.

\section{Detecting relative anomaly}\label{RelativeAnomaly}

Approaches to unsupervised anomaly detection that target the frequency criterion may not perform well in the presence of frequently occurring anomalies, as discussed in the previous sections.  We now introduce two anomaly detection models that take into account the location of the most typical observations when determining how anomalous a new observation is.  Both of these methods have the advantage that they provide a quantitative ordering of the data points in terms of how anomalous they are.  We also investigate relationships and differences with other approaches to anomaly detection, especially the approach of \cite{moonesinghe2006outlier}, which we discussed in Section \ref{vertexapproach}.

\subsection{Popularity approach}\label{popularityapproach}

We propose to consider a ``random walk'' between nodes based on the \textit{unnormalized} similarity matrix $\mathbf{S}$---instead of the transition probability matrix $\mathbf{P}$ considered in Section~\ref{vertexapproach}. 
From
\begin{equation}
\mathbf{S} =  \mathrm{diag}(\mathbf{S} \mathbf{1}) \mathbf{P},
\end{equation}
we see that the similarity $[\mathbf{S}]_{ij}$ between two nodes $\bx_i$ and $\bx_j$ factors into the transition probability $[\mathbf{P}]_{ij}$ and the vertex degree of $\bx_i$.  This has the effect that the random walk weakens when transitioning through nodes whose vertex degree is medium or small, and that it strengthens when passing through nodes of high vertex degree.  We label an observation $\bx_i$ as anomalous if its \textit{relative anomaly},
\begin{equation} \label{RA}
	\mathrm{RA}(\bx_i) \defeq - (\mathbf{s})_i,
\end{equation}
is small, where $\mathbf{s}$ is the dominant left-eigenvector of $\mathbf{S}$. This eigenvector is unique with all elements positive by the Perron--Frobenius theorem.

We can gain further insight into this algorithm via a connection with the network analysis literature.  \cite{bonacich1972factoring} considers a network of persons, where each person rates each other person as popular or not.  Their goal is to determine an overall popularity score for each person, based on the pairwise ratings.  They suggest that a measure of overall popularity of person $i$ should depend not only on how many people in the network deem that person to be popular, but also whether those people are themselves popular.  This leads to the eigenproblem $\sum_j [\mathbf{S}]_{ij} v_j = \lambda v_i$, where $v_i$ is the overall popularity of person $i$, and $[\mathbf{S}]_{ij}$ takes value one if person $i$ considers person $j$ popular. A person is labeled as overall popular when its entry in the dominant left-eigenvector of the adjacency matrix, called the eigenvector centrality, is large.  We see that by measuring anomaly using (\ref{RA}) instead of~(\ref{vddegree}), how anomalous an observation is depends not only on how many other observations are close, but also on whether these other observations themselves have close neighbors.  As a result, high vertex degree observations that are sufficiently far from many other observations in the similarity graph will be labeled anomalous.  Asymptotically, the leading eigenvector of a kernel matrix converges to the leading eigenfunction, $\varphi$, in the following eigenproblem \cite{williams2000effect}:
\begin{equation}
	\int s(\bx, \mathbf{y}) f(\bx) \varphi(\bx) \mathrm{d} \bx = \delta \varphi(\mathbf{y}).
\end{equation}
Here $f$ is the data density, and $\delta$ is the eigenvalue that corresponds to $\varphi$.  We see that, asymptotically, the popularity $\varphi(\mathbf{y})$ of an observation $\mathbf{y}$ is high if values $\bx$ that are close to $\mathbf{y}$ have high density and are popular themselves.  Here the size of the surrounding of $\mathbf{y}$ is determined by the choice of $s$.

The power method can be used to find the dominant left-eigenvector of $\mathbf{S}$.  This iterative method starts from a random initialization, $\mathbf{s}_0$, and then follows the recurrence relation $\mathbf{s}_{t+1} = \mathbf{S} \mathbf{s}_t / \Vert \mathbf{S} \mathbf{s}_t \Vert_2$. The convergence is geometric, with ratio $| \lambda_2 / \lambda_1 |$, where $\lambda_1$ and $\lambda_2$ denote the first and second dominant eigenvalue of $\mathbf{S}$, respectively.  We find that the error $\Vert\mathbf{S} \mathbf{s}_t - \mathbf{s}_t^{\mathsf{T}}\mathbf{S} \mathbf{s}_t\mathbf{s}_t\Vert_2$ typically becomes small after just a few iterations.  This computation is highly parallelizable.

We find that typically more than half of the smallest elements of the kernel matrix can be set to a small constant---allowing sparse matrix computations and a hence a speed-up of more than two---without changing the rank order of the relative anomaly values.  Furthermore, for high-dimensional problems, we can obtain good starting values for the power iteration as follows.  \cite{rahimi2007random} show that $\mathbf{S}$ can be approximated by $\boldsymbol{\Phi}^\mathsf{T} \boldsymbol{\Phi}$, where $\boldsymbol{\Phi}$ is a draw of random Fourier features calculated from the original data. If we choose only a small number of Fourier features as compared to the sample size, then rank$(\boldsymbol{\Phi}^\mathsf{T} \boldsymbol{\Phi}) \ll \mathrm{rank}(\mathbf{S})$, and we can cheaply find an approximation to the leading eigenvector of $\mathbf{S}$ as $\boldsymbol{\Phi}^\mathsf{T} \mathrm{lev}(\boldsymbol{\Phi}\boldsymbol{\Phi}^\mathsf{T})$. Here $\mathrm{lev}(\boldsymbol{\Phi}\boldsymbol{\Phi}^\mathsf{T})$ denotes the leading eigenvector of $\boldsymbol{\Phi}\boldsymbol{\Phi}^\mathsf{T}$; it can again be found using the power iteration.  In our experiments, this approach reduces the run time until the leading eigenvector of $\mathbf{S}$ is found by one fourth.

It is computationally expensive to retrain the model with every new observation.  Furthermore, it may not even be desired to update the model in the presence of every new observation, because that new observation may come from a different, anomalous data generating process. We propose to instead determine the relative anomaly of a new observation with respect to the observations in the training data set as follows.  Recall that the left-eigenproblem of $\mathbf{S}$ is $\lambda \mathbf{s} = \mathbf{S}^\mathsf{T} \mathbf{s}$, from which we see that $(\mathbf{s})_i = (\mathbf{S}^\mathsf{T} \mathbf{s})_i / \mathbf{s}^{\mathsf{T}}\mathbf{S} \mathbf{s}$.  We can use this relation to predict the relative anomaly of a new observation $\bullet$, based solely on training data, as
\begin{equation}
	\widehat{\mathrm{RA}}(\bullet)  = - \frac{(s(\bullet, \bx_1), \dots, s(\bullet, \bx_n)) \mathbf{s} }{ \mathbf{s}^{\mathsf{T}}\mathbf{S} \mathbf{s}}.
\end{equation}
This can be viewed as an application of the Nystr\"om method to approximate the leading eigenvector of the extended kernel matrix; for a reference on the Nystr\"om method, see \cite{williams2001using}.

\subsection{Shortest path approach}\label{shortpa}

We also propose an approach to relative anomaly detection based on highest similarity paths.  The idea is to first identify those observations that can be considered very typical, and then to label an observation as anomalous if it is difficult to reach it from any of the typical observations.  Here we interpret an element $[\mathbf{S}]_{ij}$ as a ``connectivity'' value between nodes $\bx_i$ and $\bx_j$.  We use the following two-step approach:
\begin{enumerate}
\item Consider those observations for which the vertex degree is higher than that of $(1-q) \cdot 100$ percent of the observations in the training data set as highly normal.  For each observation $\bullet$, we can express this as $\hat F_\mathrm{VD}(\bullet) > q$, using the empirical cumulative distribution function of vertex degrees in the training data set, $\hat F_\mathrm{VD}$. Note that by choosing the kernel bandwidth large enough we can smooth out local peaks in the data density, such that indeed the observations with highest vertex degrees can be considered normal.
\item Now, for each observation $\bullet$ that is not considered highly normal, find the length of the best-connected path from it to any of the observations deemed normal:
\begin{equation}
\max_{l\, :\, 1 - \hat F_\mathrm{VD}(\bx_l) \leq q} \
	 \max_{\left\{\substack{\mathrm{paths}\ \mathrm{from} \\ \bullet\ \mathrm{to}\ \bx_l}\right\}} \ 
	 \prod_{(i,j):\textrm{ is edge in path}} s_{ij}.
\end{equation}
Alternatively, solve the equivalent shortest path problem
\begin{align}\label{shortestpath}
	&\mathrm{RA}_q (\bullet) \defeq \\
	 &	\min_{\,l\, : 1 - \hat F_\mathrm{VD}(\bx_l) \leq q} \
	 \min_{\left\{\substack{\mathrm{paths}\ \mathrm{from} \\ \bullet\ \mathrm{to}\ \bx_l}\right\}} \ 
	 \sum_{(i,j):\textrm{ is edge in path}} - \ln s_{ij}.\notag
\end{align}
Then label $\bullet$ as anomalous if $\mathrm{RA}_q (\bullet)$ is large. $\mathrm{RA}_q (\bullet) = 1$ if $\bullet$ is one of the $q \cdot 100 $ percent of observations which are considered most normal, and $\mathrm{RA}_q (\bullet) > 1$ otherwise.  This shortest path problem can be solved more efficiently when considering a sparsified version of $\mathbf{S}$, for example by applying a directed $k$ nearest neighbor truncation. 
\end{enumerate}
An advantage of this approach is that the tuning parameter $q$ allows controlling the number of data points considered typical.  Several central regions of the data may emerge for a larger value of $q$.  A disadvantage is the higher computational complexity of the shortest path problem, which may however be reduced through subsampling.

We can gain further insight into this approach when used with the kernel function in (\ref{rbfkernel}).  Then the path length in (\ref{shortestpath}) becomes
\begin{align}
	  & \sum_{(i,j):\textrm{ is edge in path}} - \ln  \exp(-d(\bx_i,\bx_j)^2 / \gamma) \\
	 & \qquad \qquad \qquad \qquad \qquad \propto  \sum_{(i,j):\textrm{ is edge in path}} d(\bx_i,\bx_j)^2 . \notag
\end{align}
We see that the squared distance between two observations discourages large jumps, and thereby paths through high density regions are encouraged.  While the tuning parameter $\gamma$ does not influence the comparison between two path lengths, since it is only a multiplicative constant, it influences the calculation of $\hat F_\mathrm{VD}$ in (\ref{shortestpath}).  A larger value for $\gamma$ means that the bandwidth in the vertex degree estimator is higher, thereby smoothing the density more, which can be used to smear away small clusters of frequently occurring anomalies.

\subsection{Normalization}

A relative anomaly measure $\mathrm{RA}$ can be transformed into a degree of anomaly in~$(0,1)$ for each observation $\bullet$ using the empirical distribution function $\hat{F}$ of directed anomalies in the training data:
\begin{equation}
\mathrm{DORA}(\bullet) \defeq \hat{F}(\mathrm{RA}(\bullet)) \in (0,1).
\end{equation}

\subsection{Determining largest univariate deviations}

Once an anomalous state $\bx_{\mathrm{anomalous}}$ is identified, we can determine which univariate features deviate most from what is normal as follows:
\begin{enumerate}
	\item Find that normal observation in the data set which is closest to the anomalous observation:
	\begin{align}
		 &\bx_{\mathrm{closest}}(\bx_{\mathrm{anomalous}})   \\
		&\qquad\qquad = \argmin_{\bx_i :\, \mathrm{DORA}(\bx_i) < p} d(\bx_i, \bx_{\mathrm{anomalous}}).\notag
	\end{align}
	The threshold $p \in (0,1)$ determines how large the anomaly of $\bx_{\mathrm{closest}}$ may be to still be considered normal. Here it may be useful to use the $L_1$ distance to judge discrepancy, because the suggested change will be large in a few dimensions, unlike it is the case with $L_2$ distance, which will suggest smaller changes in many dimensions.  
	\item Calculate $\bx_{\mathrm{anomalous}} - \bx_{\mathrm{closest}}(\bx_{\mathrm{anomalous}})$; the largest elements of this vector difference show which univariate components need to be altered for the system to revert to a normal state.
\end{enumerate}

\section{Application} \label{Application}

We compare the relative anomaly detection approaches, introduced in Section~\ref{RelativeAnomaly}, to the vertex degree anomaly detection approach discussed in Section~\ref{vertexapproach}, using two data sets from Google, of 1,000 data points each.  We pre-process each covariate using the Box--Cox transform \cite{box1964analysis},
\begin{equation}
		x \mapsto 
		\begin{cases}
		\frac{(x + \delta)^\lambda - 1}{\lambda}, & \textrm{if } \lambda \neq 0, \\
		\ln (x + \delta), & \textrm{if } \lambda = 0,
		\end{cases}
\end{equation}
to reduce skew and normalize kurtosis;  special cases of this transform are the logarithmic and square-root transforms.  We find the parameters $(\delta, \lambda)$ as those maximizing the normal log-likelihood of the data.  We then standardize the data and form a fully connected similarity graph using the radial basis kernel.



\subsection{Potential scraping data}

The first data set contains information about potential scraping attempts.  Scraping is the automated collection of information from websites.  The two covariates are experimental features that measure aspects of user behavior for each access log.  

\begin{figure}[htbp]
\centering
\includegraphics{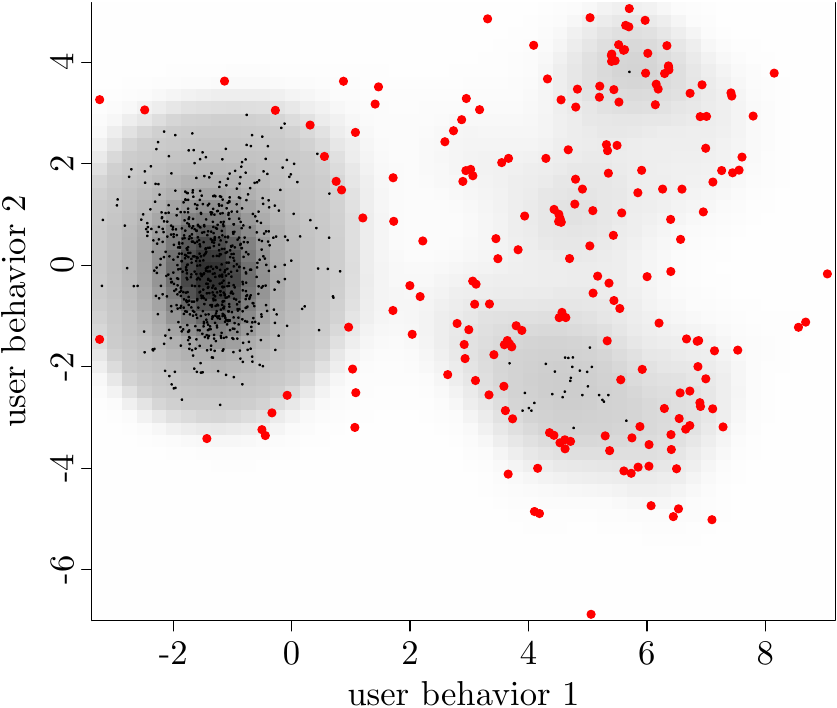}
\caption{The vertex degree approach labels low-density observation in the left cluster of normal observations as anomalous, and mistakes some observations in the diffuse right cluster of scrapers as normal; the top 20 percent detected anomalies are highlighted.}
\label{ScrapingLoc}
\end{figure}

In Figure~\ref{ScrapingLoc} we show the anomaly detection results using the vertex degree approach of Section~\ref{vertexapproach}, which targets the frequency criterion.  We set $\gamma=0.5$.  Here and in the following, the lighter the shade of grey is, the higher the respective region's detected degree of anomaly.  The top twenty percent detected anomalies are emphasized.  However, domain experts have identified that the observations in the diffuse cluster on the right exhibit behavior that is typical of scrapers.  As a result, there are false positives surrounding the very high density area around $(-1,0)$, and the observations around $(5, -2)$ and $(6,4)$ are false negatives.

\begin{figure}
\centering
\includegraphics{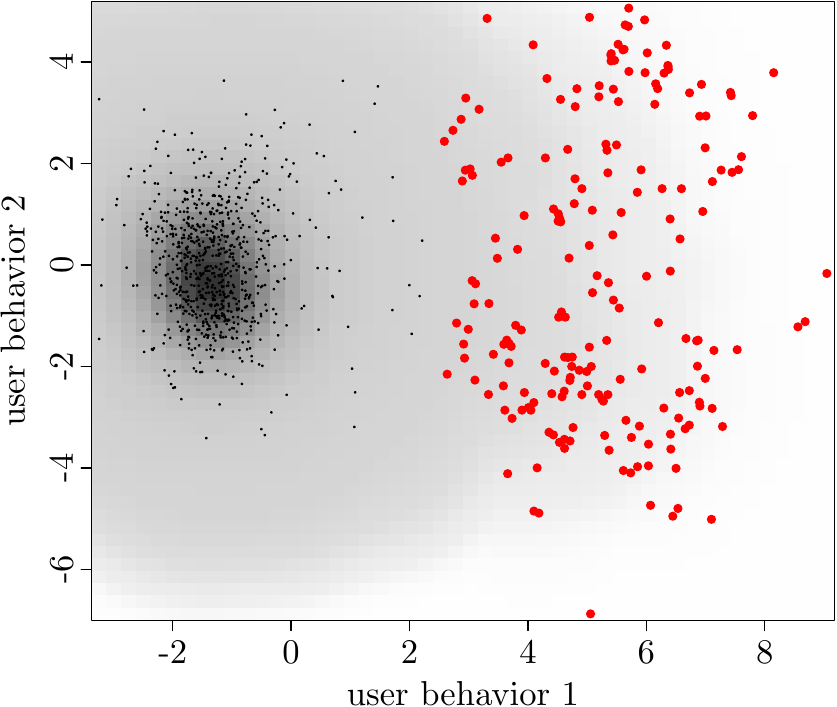}
\caption{The popularity approach correctly detects the left cluster of normal observations as normal, and labels the diffuse right cluster of scrapers as anomalous; the top 20 percent detected anomalies are highlighted.}
\label{ScrapingDirected}
\end{figure}

The results for the popularity approach to relative anomaly detection, introduced in Section~\ref{popularityapproach}, are shown in Figure~\ref{ScrapingDirected}. We set $\gamma=0.2$, because we find that the relative anomaly approach generally requires less smoothing than the vertex degree approach.  The results are not very sensitive to the exact choice of $\gamma$; lowering $\gamma$ in the vertex degree approach would result in a significant increase in the number of false positives and false negatives. There are no false positives or false negatives, as compared with the expert judgement.

It is extremely labor-intensive---potentially even impossible---to assess with certainty whether an individual data point is or is not a scraper.  Hence it may be desired to only label users as scrapers if we are very certain. 
The detected level of relative anomaly in Figure~\ref{ScrapingDirected} tends to increase while moving away from the high density area on the left. Increasing the threshold of relative anomaly above which a user is labeled as a scraper will have the desired result that only observations on the far right---whose behavior is most different from what is typical---are labeled as anomalous. In contrast, the vertex degree approach will continue labeling observations in the low density area close to the cluster of normal users as anomalous.

\begin{figure}
\centering
\includegraphics{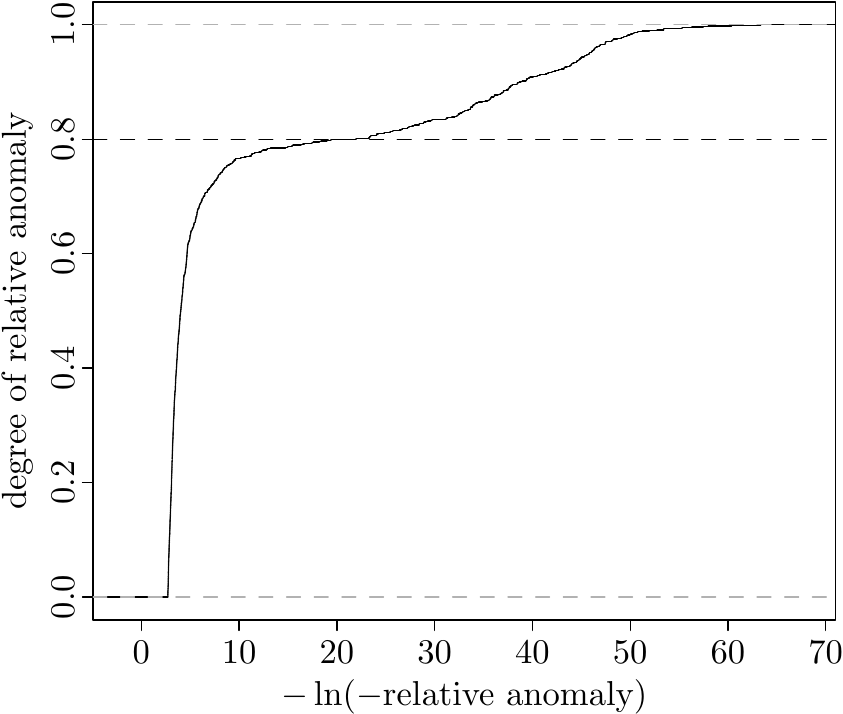}
\caption{The empirical distribution of directed anomalies can assist with deciding above which threshold of directed anomaly an observation is labeled anomalous.}
\label{VertexToAnomaly}
\end{figure}

In Figure~\ref{VertexToAnomaly} we show how the empirical cumulative distribution of relative anomalies may be useful for determining the threshold above which an observation is labeled an anomaly. For a clearer presentation, we transformed the relative anomaly values as $\bullet \mapsto - \ln(-\bullet)$. The top 20 percent of observations have much higher relative anomaly values than the other observations. This approach is particularly useful in higher-dimensional problems, where a visual inspection is difficult.

\begin{figure}
\centering
\includegraphics{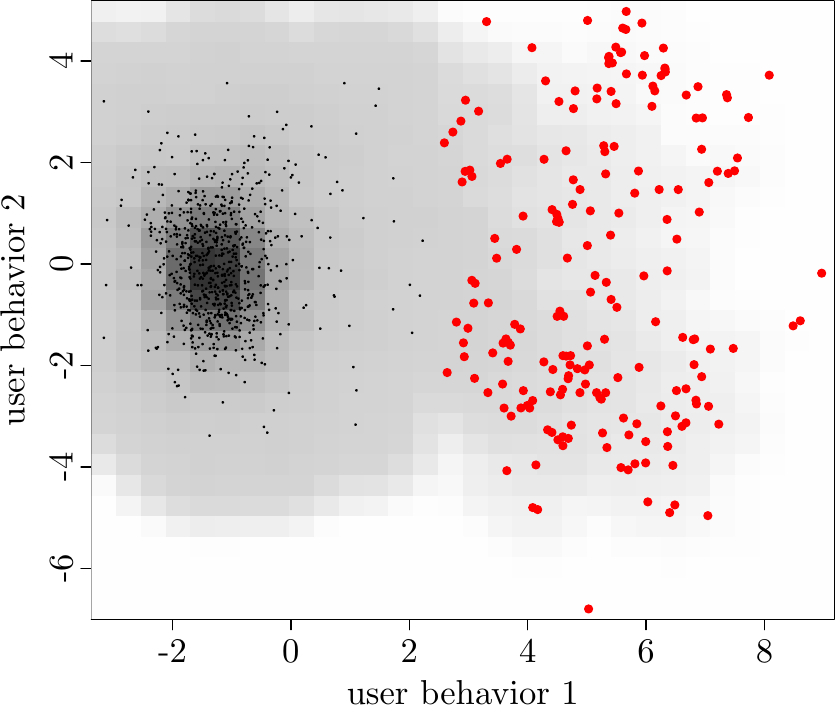}
\caption{The shortest path approach to directed anomaly detection correctly detects the left cluster of normal observations as normal, and labels the diffuse right cluster of scrapers as anomalous; the top 20 percent detected anomalies are highlighted.}
\label{pathscraping}
\end{figure}

We also apply the shortest path approach from Section \ref{shortpa} to the scraping data set. In Figure \ref{pathscraping} we see that, compared with the approach of Section \ref{RelativeAnomaly}, the shortest path approach using $q=0.5$ yields sharper bounds around the group of normal observations, which may be desired in some applications; in-sample the classification outcomes are identical.

\subsection{Wi-Fi usage data}

Our second data set contains observations on Wi-Fi channel utilization reported for wireless transmissions at different access points within a specific location in a corporate networking environment. The instantaneous channel utilization at each access point is an indication of how busy the transmission channel is, and whether the access point should change transition to a different channel. Detecting channel utilization anomalies is critical for identifying access points with low performance due to consistent high utilization. The data set contains two covariates for a Wi-Fi access point. The first covariate is measure of overall utilization, and the second covariate measures utilization of rx versus tx. 72 percent of the data points cluster at value $(-0.89, 0.04)$, which corresponds to no utilization.  According to domain experts, high utilization states are anomalous.

\begin{figure}
\centering
\includegraphics{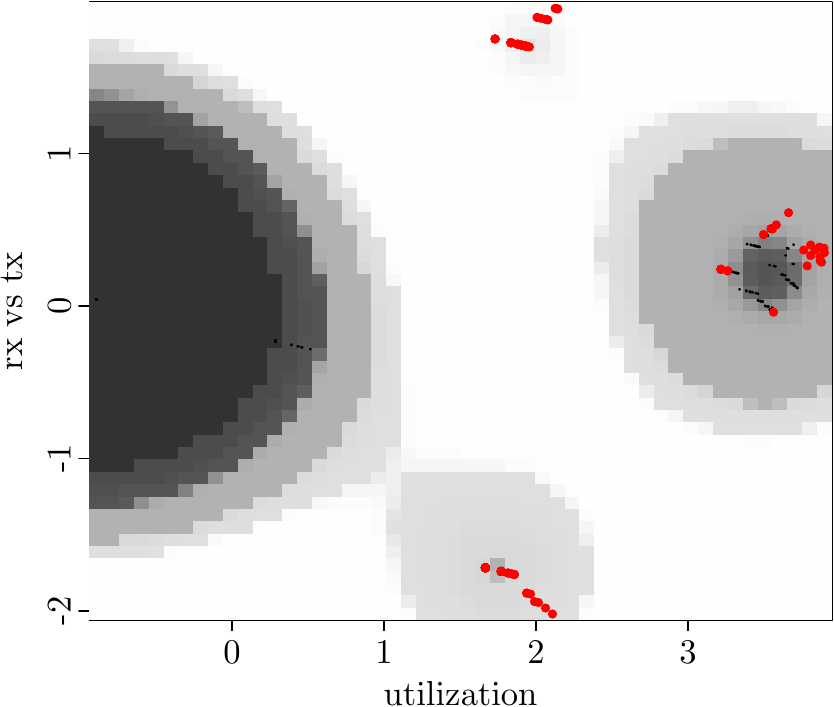}
\caption{The vertex degree approach labels the two clusters on top and bottom as anomalous, even though these correspond to medium overall Wi-Fi usage, as measured by the first covariate; most of the elements of the heavy-usage cluster on the right are labeled as normal, because heavy usage occurs relatively frequently; the top 13 percent detected anomalies are highlighted; note that a medium-usage configuration at $(2, 0)$ would falsely be considered extremely anomalous.}
\label{Wi-FiLoc}
\end{figure}

The vertex degree approach yields the results in Figure~\ref{Wi-FiLoc}, where again we set $\gamma=0.5$.  We see that the two smaller clusters around $(1.7, -1.5)$ and $(2, 1.8)$, as well as the few data points around $(0.4, -0.2)$, are jointly labeled as the top thirteen percent anomalies.  

\begin{figure}
\centering
\includegraphics{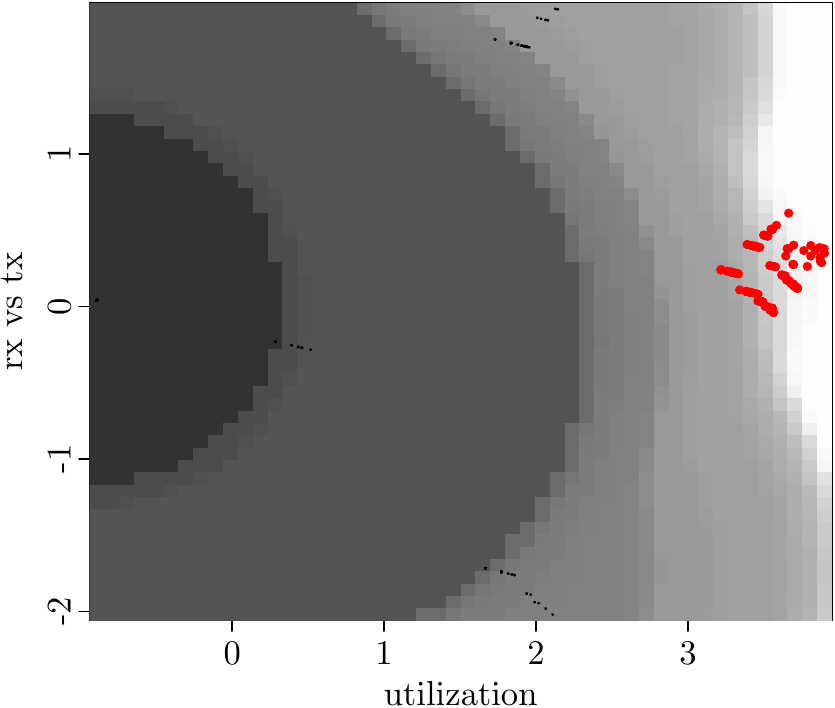}
\caption{The popularity approach correctly labels only the heavy-usage cluster on the far right as anomalous; the top 13 percent detected anomalies are highlighted; note that a medium-usage configuration at $(2, 0)$ would correctly be considered normal}
\label{Wi-FiDirected}
\end{figure}

In Figure~\ref{Wi-FiDirected} we show the results for the approach from Section \ref{popularityapproach}, again using $\gamma=0.2$.  Here the cluster of high usage observations on the far right is correctly labeled as anomalous---because it is far from the many observations at the left of the figure.  The results for the shortest path approach from Section \ref{shortpa} are similar.

%
%
%
%

\section{Conclusion}\label{Conclusion}

Unsupervised approaches to anomaly detection are commonly used because labeling data is too costly or difficult.  Many common approaches for unsupervised anomaly detection target a frequency criterion.  This means that their performance deteriorates when anomalies occur frequently, as for example in the case of scraping.  We proposed a novel concept, relative anomaly detection, that is more robust to such frequently occurring anomalies.  It is tailored to be robust towards anomalies that occur frequently, by taking into account their location relative to the most typical observations.  We presented two novel algorithms under this paradigm.  We also discussed real-time detection for new observations, and how univariate deviations from normal system behavior can be identified.  We illustrated these approaches using data on potential scraping and Wi-Fi usage from Google, Inc.

\section*{Acknowledgment}

We thank Mitch Trott, Phil Keller, Robbie Haertel and Lauren Hannah for many helpful comments, and Dave Peters as well as Taghrid Samak for granting us access to their data sets.

\ifCLASSOPTIONcaptionsoff
  \newpage
\fi



\bibliographystyle{IEEEtran}
\bibliography{library}
\end{document}